# Mediapipe and CNNs for Real-Time ASL Gesture Recognition

Rupesh Kumar
Department of CSE
Galgotias College of Engineering and Technology
Greater Noida, India
rskumar0402@gmail.com

Ashutosh Bajpai
Department of CSE
Galgotias College of Engineering and Technology
Greater Noida, India
ashutoshbajpai161@gmail.com

Ayush Sinha
Department of CSE
Galgotias College of Engineering and Technology
Greater Noida, India
ayushsinha132001@gmail.com

S.K Singh
Professor Department of CSE
Galgotias College of Engineering and Technology
Greater Noida, India
Drshashikant.singh@galgotiacollege.edu

*Abstract*— **This research paper describes a real-time system for identifying American Sign Language (ASL) movements that employs modern computer vision and machine learning approaches. The suggested method makes use of the Mediapipe library for feature extraction and a Convolutional Neural Network (CNN) for ASL gesture classification.**

**The testing results show that the suggested system can detect all ASL alphabets with an accuracy of 99.95%, indicating its potential for use in communication devices for people with hearing impairments. The proposed approach can also be applied to additional sign languages with similar hand motions, potentially increasing the quality of life for people with hearing loss.**

**Overall, the study demonstrates the effectiveness of using Mediapipe and CNN for real-time sign language recognition, making a significant contribution to the field of computer vision and machine learning.**

## I. INTRODUCTION

Sign language serves as a vital means of communication for the hearing and speech-impaired community, relying on hand gestures, facial expressions, and body language. While various sign languages are used globally, the development of automatic sign language recognition (SLR) systems becomes crucial to facilitate communication between sign language users and those without hearing loss. These systems analyse multimodal data, including hand and body motions, facial expressions, and overall movement, to create assistive technology for the deaf and hard of hearing.

SLR encompasses three main subproblems: static, isolated, and continuous recognition. Static recognition focuses on identifying finger-spelled gestures representing alphabets and digits, while isolated recognition involves recognizing dynamic gestures corresponding to individual words. Continuous recognition tackles the challenge of recognizing dynamic gesture sequences that encompass phrases combined with non-sign portions.

Different approaches exist for SLR, depending on the employed techniques. Two commonly used methods are glove-based and vision-based approaches. Vision-based approaches rely on image and signal processing techniques, analysing pictures or videos as input to determine the relevant sign. Glove-based methods, on the other hand, utilize data gloves to capture hand movements and positions. Each approach presents its own advantages and limitations, including computational costs, accuracy standards, and equipment requirements.

The continuous recognition of sign-language using a convolutional neural network (CNN) and data augmentation approaches will be the main emphasis of this research article. We will examine the benefits and drawbacks of this strategy and assess its effectiveness using data that is readily available to the public. Our goal is to aid in the creation of more precise and effective sign language recognition systems that will increase accessibility and communication for the deaf community.

## II. PREVIOUS WORK

The global deaf community heavily depends on sign language as their primary mode of communication. As computer vision and machine learning continue to advance, researchers are actively developing computerized systems capable of recognizing and converting sign language gestures into text or spoken words. This article delves into the latest research surrounding machine learning-based SLR, aiming to provide insights into the ongoing advancements in this field.

Das et al. [1] created a deep learning-based SLR system employing processed static images of ASL motions. They attained an average accuracy rate of more than 90% by training an Inception V3 CNN on a dataset of 24 classes representing alphabets from A to Z, except for J, with the best validation accuracy reaching 98%. When given correctly cropped image datasets, the researchers determined that the Inception V3 model is sufficient for static sign language detection.

A. K. Sahoo [2] focused on identifying Indian sign language (ISL) using machine learning techniques. Their study specifically targeted static hand movements corresponding to numbers 0 to 9. By utilizing a digital RGB sensor to capture images of the signs, the researchers built a dataset consisting of 500 photos, with one image per digit. They trained models using supervised learning approaches like Naive Bayes and k-Nearest Neighbor, achieving average accuracy rates of 98.36% and 97.79%, respectively, with k-Nearest Neighbor slightly outperforming Naive Bayes.

Ansari et al. [3] investigated the classification of static movements in ISL using images incorporating 3D depth data. They employed Microsoft Kinect to capture both 3D depth data and 2D images. The dataset comprised 5041 static hand gesture photos classified into 140 classes. The model was trained using K-means clustering, which resulted in an average accuracy rate of 90.68% for recognising 16 alphabets.

Rekha et al. [4] analysed a dataset containing 23 static and three dynamic signs in ISL. They employed skin color segmentation techniques to detect hands and trained a multiclass Support Vector Machine (SVM) using features such as edge orientation and texture. The SVM achieved a success rate of 86.3%. However, due to its slow processing speed, this method was not suitable for real-time gesture detection.

Bhuyan et al. [5] utilized a dataset of 400 photos representing eight motions from ISL. They adopted a skin color-based segmentation approach to detect hands and employed the nearest neighbor classification method, achieving a recognition rate of over 90%.

Pugeault et al. [6] developed a real-time recognition system for ASL alphabets using a dataset of 48,000 3D depth photos collected through a Kinect sensor. They achieved highly accurate classification rates by incorporating Gabor filters and multi-class random forests.

Keskin et al. [7] recognized ASL numerals by employing a technique based on object identification using components. Their dataset consisted of 30,000 observations categorized into ten classes.

Sundar B et al. [8] presented a vision-based approach for recognizing ASL alphabets using the MediaPipe framework. Their system achieved an accuracy of 99% in recognizing 26 ASL alphabets through hand gesture recognition using Long Short-Term Memory (LSTM). The proposed approach can convert hand gestures into text, making it valuable for human-computer interaction (HCI). The combination of MediaPipe hand landmarks and LSTM proved effective in gesture recognition for HCI applications.

Jyotishman Bora et al. [9] developed a machine learning approach to recognize Assamese Sign Language. They used a combination of 2D and 3D images and MediaPipe hand tracking solution to train a feed-forward neural network. The model achieved 99% accuracy in recognizing Assamese gestures. The study highlights the effectiveness of their method for other alphabets and gestures in the language and suggests its applicability to other local Indian languages. The MediaPipe solution provides accurate tracking and faster classification, while its lightweight nature allows implementation on different devices without compromising speed and accuracy.

Arpita Halder et al. [10] introduced a simplified SLR methodology using MediaPipe's framework and machine learning algorithms. Their model achieved an average accuracy of 99% in multiple sign-language datasets, enabling real-time and precise detection without the need for wearable sensors. The approach offers a lightweight and cost-effective solution, surpassing complex, and computationally intensive methods. The study showcases MediaPipe's efficiency and adaptability to regional sign languages. Here is a comparison table of all the studies that were used in current iterations of this study.

TABLE I: COMPARISON TABLE FOR SIGN LANGUAGE DETECTION TECHNIQUE

| Paper | Techniques Used | Accuracy Achieved | Dataset |
| --- | --- | --- | --- |
| Das et al. [1] | CNN- Inception V3 | >90% | ASL[a] |
| A. K. Sahoo [2] | Hierarchical centroid feature vector. Naive Bayes and KNN. | KNN- 98.36%, Naive Bayes- 97.79% | ISL[b] |
| Ansari et al. [3] | Microsoft Kinect Camera, VFH[c], SIFT[d], and SURF[e], nearest neighbour (K-d tree) | 90.68% | ISL[b] (Word Based) |
| Rekha et al. [4] | KNN, SVM | 89%, 91% | User-generated dataset |
| Bhuyan et al. [5] | Geometric modeling and texture analysis | IM1 93.5%, IM2 92.0% TIM1 95.5%, TIM2 93.5% IMR1 95.0%, IMR2 91.5% IMRL1 94.5%, IMRL2 91.5% | ISL[b] |
| Pugeault et al. [6] | Kinect sensor and multi-class random forest | 75% | ASL[a] alphabets |
| Keskin et al. [7] | Kinetic-depth sensor, Random-forest, SVM | 99% on live depth images in real-time | Ten digits of ASL[a] |
| Sundar B et al. [8] | Mediapipe and LSTM as image classifier | 99% | User-generated ASL[a] alphabets |
| Jyotishman Bora et al. [9] | Mediapipe and sign classification with Deep Learning | 99% | User-generated Assamese gestures |
| Arpita Halder et al. [10] | Mediapipe with SVM | 99% | Multiple datasets such as American, Indian, Italian and Turkey |

[a.] American Sign-Language
[b.] Indian Sign-Language
[c] Viewpoint Feature Histogram
[d] Scale-Invariant Feature Transform
[e] Speeded Up Robust Features

In conclusion, SLR is an evolving field of study that has the potential to significantly enhance communication for millions of individuals worldwide. Researchers have made notable progress in developing SLR systems that can effectively translate gestures into text or voice using computer vision and machine learning approaches. The studies mentioned in this post demonstrate the effectiveness of classifiers like Naive Bayes and k-Nearest Neighbor, as well as deep learning models such as Inception V3 and K-means clustering, in recognizing both static and dynamic sign language gestures with high accuracy rates. However, there is still much work to be done in creating more robust and efficient systems that can handle a wider range of sign language motions and variations in lighting, camera angles, and backgrounds. These developments have the potential to significantly enhance the lives of those who use sign language as their primary way of communication.

The ongoing research in SLR highlights the importance of employing computer vision and machine learning techniques to bridge the communication gap for the deaf community. These advancements have the potential to enhance accessibility and promote inclusive communication on a global scale.

### III. PROPOSED ARCHITECTURE

Our proposed architecture for SLR aims to accurately interpret and classify ASL gestures. To achieve this, we employ a multi-step process that involves image frame acquisition, hand tracking, feature extraction, and classification. By leveraging a large ASL dataset and state-of-the-art techniques, our architecture enables the model to capture the intricate details and movements of ASL gestures with precision. Figure 2.1 illustrates the overall flow of our proposed SLR architecture. This diagram provides a visual representation of the sequential steps involved in our system.

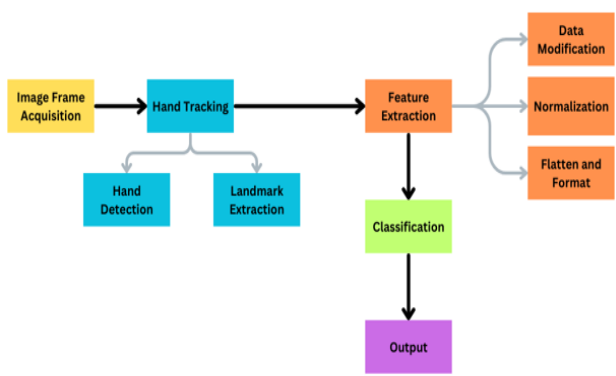

Fig. 3.1 Workflow for ASL model

#### A. Image Frame Acquisition

To develop an accurate Sign Language Recognition (SLR) model, it is crucial to acquire high-quality image frames that capture the gestures and movements of American Sign Language (ASL). In our proposed architecture, we utilize the ASL dataset, which consists of a diverse range of ASL gestures performed by individuals proficient in ASL. Our dataset encompasses 26 distinct classes corresponding to the alphabets from A to Z. By incorporating these 26 classes into our ASL dataset, we ensure that the model can accurately recognize and classify a wide range of ASL gestures. Each class is represented by a substantial number of images, with 4500 samples per class. This large dataset size enables our proposed architecture to leverage a significant amount of data during the training process. Consequently, the model can effectively learn the intricate hand movements and subtle variations associated with ASL gestures. Utilizing a dataset of this magnitude allows our SLR model to capture both subtle nuances and distinct characteristics of ASL gestures with high precision.

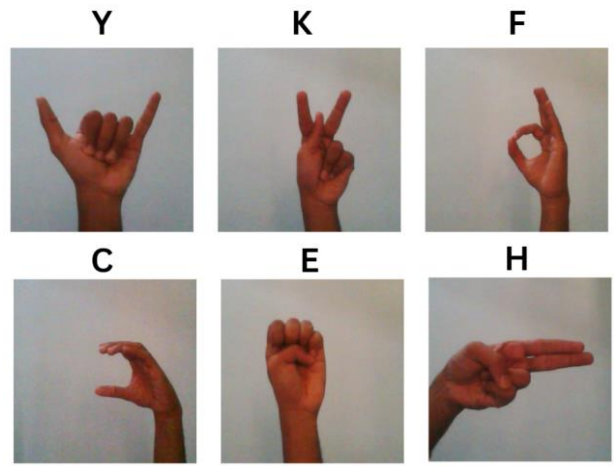

Fig. 3.2 ASL Dataset used for model training

#### B. Hand Tracking

In our proposed architecture for SLR we leverage the Mediapipe module, an open-source project developed by Google, to perform accurate hand tracking. The Mediapipe module offers robust and efficient hand pose estimation, enabling us to track the movements and positions of both hands in real-time. From the hand tracking module, we extract a total of 21 landmarks for each hand, capturing their spatial configuration and movements. These landmarks serve as essential features for subsequent stages of the sign language recognition model.

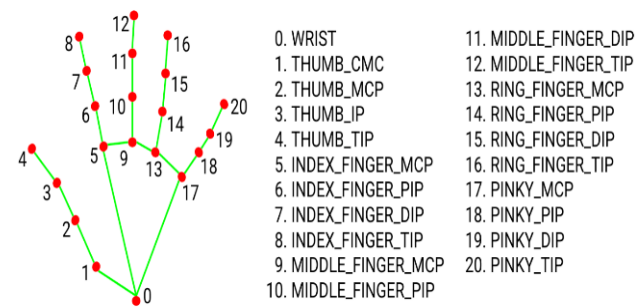

Fig. 3.3 Landmarks from Mediapipe Hand Tracking Module

#### C. Feature Extraction

The following steps were taken to extract frames and prepare the appropriate data from the 21 landmarks for classification:

a) *Data Modification:* The hand landmark coordinates were adjusted relative to the hand's centre by subtracting the centre point coordinates from each landmark coordinate. The modified landmark coordinates were then translated to have a non-negative value for all fields, ensuring that the hand's relative spatial information was preserved.

b) *Normalization*: The translated landmark coordinates were normalized by dividing them by a scaling factor, derived from the hand bounding box dimensions or another suitable metric. This step ensured consistent normalization across different hand sizes.

c) *Flatten and Format:* The normalized landmark coordinates were flattened into a 1D array by concatenating the space-coordinates of each landmark. The resulting array was formatted appropriately for classification. The resulting 1D array will serve as the feature representation for the hand gesture in the frame. This feature vector can be fed into a classification algorithm to recognize and classify the corresponding sign language gesture.

*D. Classification*

In our proposed architecture, we utilize a multi-layer neural network to effectively classify sign language gestures. This neural network takes a feature vector obtained from the input and predicts the corresponding sign language motion. The architecture consists of interconnected layers, including input, hidden, and output layers. These layers are composed of numerous neurons that process the incoming data to generate an output. Throughout the training process, the network's parameters, such as weights and biases, are iteratively learned over 50 epochs to enhance the accuracy of the classification.

The CNN architecture is applied in the model to effectively process and analyse the input features derived from the landmark coordinates, leveraging its capabilities in extracting and learning meaningful patterns from non-visual data as well.. With 42 inputs in its initial input layer, the CNN analyses the features extracted from the sign language gestures. The output layer of the network produces a classification result within a set of predefined classes, ranging from 'A' to 'Z'. This comprehensive architecture, trained over multiple epochs, demonstrates its potential for accurate sign language gesture recognition.

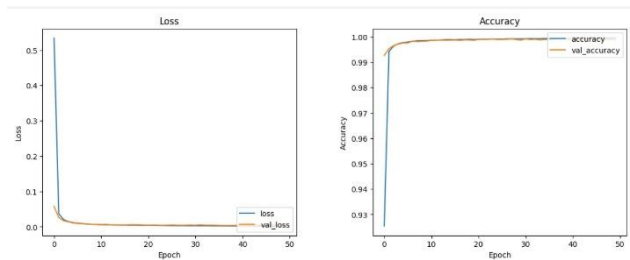

Fig. 3.4 Loss and Accuracy Graph during Model Training

*E. Output Gesture*

In the classification phase of our proposed architecture, the goal is to predict and return the corresponding sign language gesture as a value between 'A' and 'Z'. The output gesture represents the recognized sign language letter based on the input features and the trained model. After the feature vector is passed through the neural network, the final layer of the network produces a probability distribution over different classes or gesture labels. Each class corresponds to a specific sign language letter. To obtain the predicted gesture, we select the class with the highest probability. The output gesture is then determined by mapping the selected class to the corresponding letter between 'A' and 'Z'.

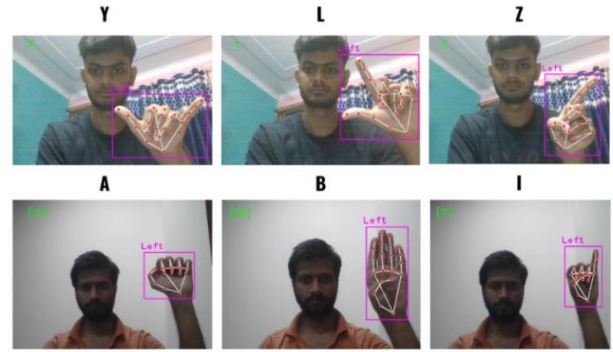

Fig. 3.5 Output of working ASL-model

## IV. RESULT ANALYSIS

We cleaned the ASL dataset before using 4500 photos per class to train our model. There were 166K photos in the original collection. An 80% training set and a 20% test set were created from the dataset. In order to train the model, we used a range of hyperparameters, including learning rate, batch size, and the number of epochs.

Our test set evaluation metrics demonstrate the trained model's remarkable performance. It properly identified every sample in the test set, earning a high accuracy score of 100%. The classification report's precision, recall, and F1-score values are all 100%, showing that the model properly identified each class's samples without making any errors.

The F1 score is a metric that combines precision and recall to provide a single measure of performance. Precision measures the accuracy of identifying positive instances, while recall measures the ability to capture all positive instances. The F1 score is calculated using the harmonic mean of precision and recall, as shown in the formula:

$$F1\ Score\ = \frac{2 * (Precision * Recall)}{(Precision + Recall)} \quad (1)$$

This score offers a balanced evaluation ranging from 0 to 1, where higher values indicate better performance in both precision and recall.

TABLE II: CLASSIFICATION REPORT FOR ASL-MODEL

| Classes | Precision | Recall | F1-score | Support |
|---|---|---|---|---|
| A | 1.00 | 1.00 | 1.00 | 912 |
| B | 1.00 | 1.00 | 1.00 | 940 |
| C | 1.00 | 1.00 | 1.00 | 921 |
| D | 1.00 | 0.99 | 1.00 | 927 |
| E | 1.00 | 1.00 | 1.00 | 900 |
| F | 1.00 | 1.00 | 1.00 | 923 |
| G | 1.00 | 1.00 | 1.00 | 910 |
| H | 1.00 | 1.00 | 1.00 | 895 |
| I | 1.00 | 1.00 | 1.00 | 884 |
| J | 1.00 | 1.00 | 1.00 | 874 |
| K | 1.00 | 1.00 | 1.00 | 868 |
| L | 1.00 | 1.00 | 1.00 | 893 |
| M | 0.99 | 1.00 | 0.99 | 884 |
| N | 1.00 | 0.99 | 1.00 | 935 |
| O | 1.00 | 1.00 | 1.00 | 887 |
| P | 1.00 | 1.00 | 1.00 | 898 |
| Q | 0.99 | 1.00 | 1.00 | 837 |
| R | 1.00 | 1.00 | 1.00 | 912 |

| S | 1.00 | 1.00 | 1.00 | 861 |
|---|---|---|---|---|
| T | 1.00 | 1.00 | 1.00 | 895 |
| U | 1.00 | 1.00 | 1.00 | 873 |
| V | 1.00 | 1.00 | 1.00 | 901 |
| W | 1.00 | 1.00 | 1.00 | 917 |
| X | 1.00 | 1.00 | 1.00 | 952 |
| Y | 1.00 | 1.00 | 1.00 | 897 |
| Z | 1.00 | 1.00 | 1.00 | 904 |
| Accuracy | | | 1.00 | 23400 |
| Macro avg | 1.00 | 1.00 | 1.00 | 23400 |
| Weighted avg | 1.00 | 1.00 | 1.00 | 23400 |

The confusion matrix provides a summary of the performance of a classification model. Each row in the matrix represents the instances in the actual class, while each column represents the instances in the predicted class. Fig 3.7 represents the confusion matrix plotted between the 26 classes representing the alphabets (A-Z).

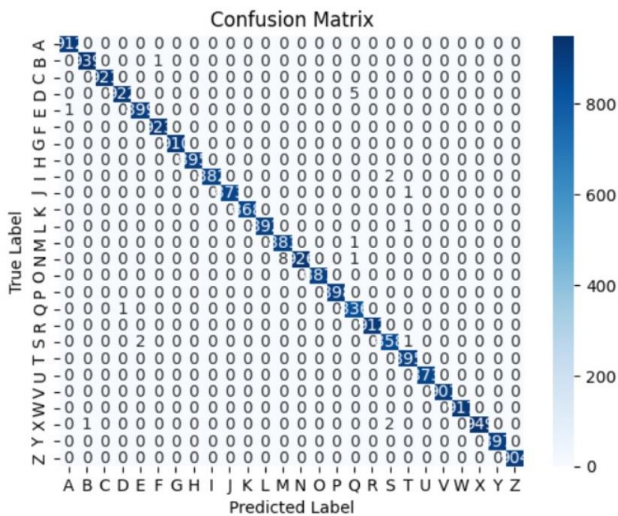

Fig. 4.1 Confusion matrix

## V. CONCLUSION AND FUTURE SCOPE

Convolutional neural network (CNN) classification and feature extraction were used to construct an effective American Sign Language recognition model. With a 99.95% accuracy score on the test set and 100% precision, recall, and F1-score values for all classes, the model performed exceptionally well. Data augmentation techniques were employed to increase variety and were added to the dataset that was used for training and testing.

To increase the model's accuracy and speed, future research and development in this field may examine other deep learning architectures and approaches. The model's applicability and inclusivity could be further increased by adding support for more sign languages and gestures. Making use of sign language recognition technology to create a bidirectional communication application is another interesting future application for this technology. The accessibility and inclusivity of sign language users would be considerably improved by such an application, especially when they are interacting with non-sign language users.

In conclusion, the creation of an ASL recognition model is a significant accomplishment in the field of sign language recognition and has the potential to significantly improve sign language users' accessibility and communication. The development of more comprehensive SLR systems and bidirectional communication applications employing sign language recognition technology may result from additional research and development in this field. More study and development could lead to a broader application of this technology in the real world, expanding accessibility and communication for sign language users and raising their quality of life.

.